\documentclass[runningheads]{llncs}

\usepackage{eccv}

\usepackage{eccvabbrv}

\usepackage{graphicx}
\usepackage{booktabs}
\usepackage{amssymb}
\usepackage{bbm}
\usepackage{multirow}
\usepackage{float}
\usepackage{soul}
\usepackage{subcaption}

\usepackage[accsupp]{axessibility}  

\usepackage[pagebackref,breaklinks,colorlinks,citecolor=eccvblue]{hyperref}

\usepackage{orcidlink}

\begin{document}

\title{CIFA: Contextual-Intersectional Fairness Auditing for Hidden Subgroup Discovery in Face Analysis}

\titlerunning{CIFA: Contextual-Intersectional Fairness Audit}

\author{
Nazia Aslam\inst{1,2}\orcidlink{0000-0002-8381-9702}
\and 
Khalid Adnan Alsayed\inst{3}\orcidlink{0009-0000-6272-6761}
\and 
Thomas B. Moeslund\inst{1}\orcidlink{0000-0001-7584-5209}
\and 
Kamal Nasrollahi\inst{1,4} 
}

\authorrunning{Aslam \etal}

\institute{
Aalborg University, Denmark \\ \email{\{naas, kn, tbm\}@create.aau.dk}
\and Pioneer Centre for AI, Denmark 
\and 
Ducaltus Ltd., Newcastle upon Tyne, United Kingdom\\ \email{research@ducaltus.com} 
\and 
Milestone Systems, Denmark
}

\maketitle

\begin{abstract}
Fairness evaluation in computer vision commonly relies on aggregate accuracy and demographic subgroup analysis. However, visual models are also sensitive to contextual factors such as illumination, blur, image quality, facial accessories, and appearance attributes. These factors may interact with demographic characteristics, producing hidden subgroups in which performance degrades substantially despite strong aggregate accuracy and apparently acceptable demographic fairness. To address this, we propose the Contextual-Intersectional Fairness Auditing Framework (CIFA), a structured framework for identifying subgroup vulnerabilities arising from interactions between demographic and contextual attributes. CIFA performs demographic, contextual, and contextual-intersectional auditing, followed by worst-group discovery to identify and rank the most vulnerable attribute combinations. We evaluate CIFA on gender classification using ResNet-50 \cite{he2016deep} and ViT-B/16 \cite{dosovitskiy2020image} across FairFace \cite{Karkkainen2021}, CelebA \cite{Liu2015}, and UTKFace \cite{Zhang2017}. Our results show that aggregate accuracy and demographic-only evaluation can mask substantial contextual-intersectional disparities. We further assess several established mitigation strategies through an audit--mitigate--reaudit protocol and find that, although some worst-group disparities are reduced, no single strategy consistently eliminates them across datasets and architectures. These findings establish contextual-intersectional auditing as an important component of fairness evaluation and provide a reproducible framework for discovering, prioritizing, and reassessing hidden subgroup risks in face analysis systems. 

\keywords{Fairness Auditing \and Face Analysis \and Contextual Bias \and Intersectional Fairness \and Hidden Subgroup Failure}
\end{abstract}

\section{Introduction}
\label{sec:intro}

\begin{figure}[]
    \centering
    \begin{subfigure}[t]{0.28\linewidth}
        \centering
        \includegraphics[width=3cm,height=3.5cm]{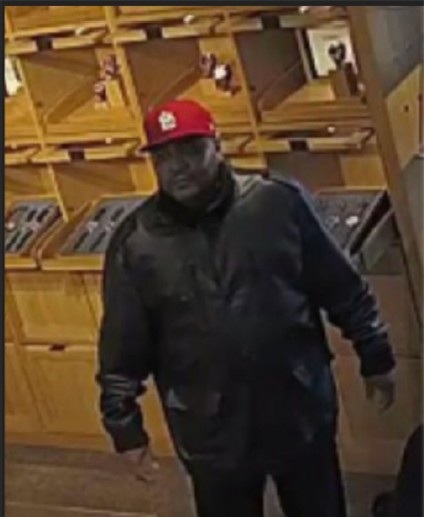}
        \caption{}
        \label{fig:demographic_audit}
    \end{subfigure}
    \begin{subfigure}[t]{0.28\linewidth}
        \centering
        \includegraphics[width=3cm,height=3.5cm]{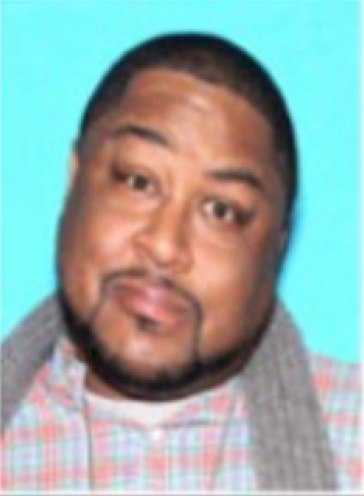}
        \caption{}
        \label{fig:contextual_intersectional_audit}
    \end{subfigure}
    \caption{Real-world facial recognition failure. Left: crime-scene surveillance image. Right: Robert Williams, who was incorrectly matched to the surveillance image by an automated facial recognition system. \cite{burtonharris2020wrongfully}}
    \label{fig:Face_recognition_system}
\end{figure}

Computer vision systems are increasingly deployed in high-stakes applications, including facial recognition \cite{li2020review}, identity verification \cite{han2025continuous}, surveillance \cite{aslam2022a3n}, video anomaly detection\cite{aslam2022unsupervised,aslam2024demaae,aslam2024transganomaly},  healthcare \cite{aslam2022attention}, and automated decision-support systems \cite{nadibaidze2024ai}. As these technologies become embedded in socially consequential settings, concerns regarding fairness, bias, and equitable performance have emerged as critical challenges \cite{Buolamwini2018}. The real-world consequences of such biases were highlighted by the wrongful arrest of Robert Williams, an innocent Black man who was misidentified by a facial recognition system \cite{burtonharris2020wrongfully}. While many vision systems achieve high overall accuracy, their performance often degrades when contextual factors interact with demographic characteristics. This limitation largely stems from training datasets that insufficiently capture both demographic diversity and the range of real-world contextual conditions. These effects can combine to produce contextual-intersectional bias, where underrepresented groups, particularly Black women and Asian populations, experience disproportionately higher error rates in challenging contexts. Current fairness evaluations primarily focuse on demographic subgroup analysis based on attributes such as gender, race, ethnicity, and age \cite{Barocas2023}, using metrics such as demographic parity, equal opportunity, and equalized odds \cite{Hardt2016}. However, these evaluations largely overlook the interaction between demographic and contextual factors. As a result, models that appear fair under demographic-only assessment may still exhibit significant failures within specific contextual-intersectional subgroups \cite{Geirhos2020,Beery2018,OakdenRayner2020}. Despite extensive research on demographic fairness, intersectionality, dataset bias, and worst-group robustness \cite{Kearns2018,Sagawa2020}, limited attention has been given to jointly auditing contextual and demographic factors within a unified framework.

To address this gap, we propose the Contextual-Intersectional Fairness Auditing Framework (CIFA), a framework that systematically evaluates fairness across demographic, contextual, and contextual-intersectional dimensions. CIFA performs demographic, contextual, and contextual-intersectional auditing, followed by systematic worst-group discovery to identify hidden subgroup vulnerabilities that remain undetected under conventional evaluation. Furthermore, it provides a foundation for an audit--mitigate--re-audit pipeline, enabling the measurement of improvements in subgroup fairness, contextual robustness, and worst-group performance. We hypothesize that systematic contextual-intersectional auditing reveals hidden subgroup vulnerabilities that remain obscured under aggregate and demographic-only evaluation. Unlike prior work that evaluates fairness primarily across demographic groups \cite{Geirhos2020,Beery2018,OakdenRayner2020}, CIFA provides a unified audit--mitigate--re-audit framework that jointly evaluates demographic, contextual, and contextual-intersectional vulnerabilities while systematically verifying mitigation effectiveness. The contributions of this work are as follows:

\begin{enumerate}
    \item We propose CIFA, a contextual-intersectional fairness auditing framework that evaluates models across demographic, contextual, and contextual intersectional subgroups, with worst-group discovery for identifying hidden failure modes.
    
    \item We conduct cross-dataset and cross-architecture experiments on FairFace \cite{Karkkainen2021}, CelebA \cite{Liu2015}, and UTKFace \cite{Zhang2017}, showing that contextual-intersectional disparities persist across datasets.

    \item We show that high aggregate accuracy can conceal substantial subgroup failures, with worst group accuracy gaps of up to 26.43\%, and evaluate an audit--mitigate--re-audit pipeline to measure whether mitigation strategies reduce these disparities.
\end{enumerate}

\section{Related Work}

Fairness in computer vision has been studied extensively, particularly in face analysis \cite{li2020review} and identity-related tasks \cite{han2025continuous}. Buolamwini and Gebru showed that commercial gender classification systems exhibit unequal performance across demographic groups, with darker-skinned women experiencing the highest error rates \cite{Buolamwini2018}. Their study highlighted the importance of evaluating model behavior across overlapping demographic groups rather than treating each attribute independently. Raji \etal further demonstrated that independent audits can expose such disparities in deployed AI systems and support greater accountability in their development and use \cite{Raji2019}. More recent studies indicate that demographic disparities remain a concern not only for conventional computer vision models, but also for modern foundation models \cite{Gustafson2023,Dehdashtian2024}. While balanced demographic representation is an important step toward fairer evaluation, it does not guarantee that all sources of model failure are captured. In particular, demographic balance alone may not reveal vulnerabilities that arise when demographic attributes interact with visual context.

Intersectional fairness provides a natural motivation for moving beyond single-attribute evaluation. It argues that harms may be missed when protected attributes are evaluated independently \cite{Crenshaw1989}. Fairness gerrymandering formalizes this concern by showing that models can satisfy fairness constraints on coarse demographic groups while still violating them on finer-grained subgroups \cite{Kearns2018}. Existing intersectional fairness work has primarily focused on demographic intersections, such as race $\times$ gender or age $\times$ gender. CIFA extends this perspective by incorporating contextual attributes into the auditing process. Rather than auditing demographic attributes alone, we evaluate whether visual factors such as illumination, image quality, pose, occlusion, and background interact with demographic attributes to produce hidden subgroup failures.

This perspective is also connected to work on dataset bias, spurious correlations, and worst-group robustness. Dataset bias has long been recognized as a limitation of computer vision evaluation, with Torralba and Efros showing that recognition datasets contain dataset-specific biases that affect cross-dataset generalization \cite{Torralba2011}. Related work on shortcut learning shows that deep neural networks can achieve strong aggregate performance by exploiting spurious correlations rather than learning robust task-relevant representations \cite{Geirhos2020}. In visual recognition, such correlations often arise from contextual cues, including background, illumination, image quality, and capture conditions. Group robustness benchmarks further show that models can under perform on minority groups when labels are correlated with contextual attributes \cite{Sagawa2020}. These findings suggest that aggregate accuracy may obscure systematic failures caused by interactions among labels, demographic attributes, and visual context.

A related issue is hidden stratification, in which high overall performance conceals weak performance on important subgroups. Oakden-Rayner \etal describe this problem in medical imaging, showing that models can fail on clinically important subsets despite high aggregate accuracy \cite{OakdenRayner2020}. Worst-group robustness methods address this issue by optimizing for the most poorly performing group rather than average performance alone, with distributionally robust optimization commonly used to improve robustness under group shifts and reduce worst-group error \cite{Sagawa2020}. However, these methods depend on meaningful subgroup definitions. Algorithmic auditing provides a complementary perspective by systematically evaluating models to identify disparities before or during deployment. Prior auditing work has focused mainly on demographic disparities, failure documentation, and accountability in commercial systems \cite{Raji2019}. Recent work further emphasizes the need for structured and reproducible audit protocols, particularly in high-stakes settings \cite{Lacmanovic2025}. CIFA addresses this gap by jointly auditing demographic, contextual, and contextual-intersectional factors, enabling hidden subgroup discovery and supporting an audit--mitigate--re-audit workflow in which identified vulnerabilities can be targeted and reassessed under the same evaluation protocol.

\subsection{Positioning of CIFA}
\label{subsec:cifa_positioning}

Prior work addresses complementary aspects of fairness evaluation. Gender Shades and FACET establish demographic and multi-attribute auditing for visual models \cite{Buolamwini2018,Gustafson2023}; subgroup-fairness methods search for violations across structured protected-attribute groups \cite{Kearns2018}; AIF360 and Fairlearn provide general-purpose assessment and mitigation tools \cite{Bellamy2019,Bird2020}; and Group DRO improves performance over predefined worst-case groups \cite{Sagawa2020}. CIFA complements these directions by treating visual context as an explicit auditing dimension and by preserving the same subgroup definitions throughout auditing, mitigation evaluation, and re-auditing.

\begin{table}[]
\centering
\caption{Comparison of representative fairness evaluation methods.}
\label{tab:framework_comparison}
\small
\setlength{\tabcolsep}{5pt}
\resizebox{\linewidth}{!}{
\begin{tabular}{ccccccc}
\toprule
\textbf{Method} &
\begin{tabular}[c]{@{}c@{}}\textbf{Demographic}\\ \textbf{Audit}\end{tabular} &
\begin{tabular}[c]{@{}c@{}}\textbf{Contextual}\\ \textbf{Audit}\end{tabular} &
\begin{tabular}[c]{@{}c@{}}\textbf{Contextual-}\\ \textbf{Intersectional}\\ \textbf{Audit}\end{tabular} &
\begin{tabular}[c]{@{}c@{}}\textbf{Hidden}\\ \textbf{subgroup}\\ \textbf{discovery}\end{tabular} &
\textbf{Mitigation} &
\textbf{Re-Audit} \\
\midrule

\begin{tabular}[c]{@{}c@{}}
Gender\\ Shades \cite{Buolamwini2018}
\end{tabular}
& $\checkmark$ & $\times$ & $\times$ & $\times$ & $\times$ & $\times$ \\

AI Fairness 360 \cite{Bellamy2019}
& $\checkmark$ & $\times$ & $\times$ & $\times$ & $\checkmark$ & $\times$ \\

Fairlearn \cite{Bird2020}
& $\checkmark$ & $\times$ & $\times$ & $\times$ & $\checkmark$ & $\times$ \\

FACET \cite{Gustafson2023}
& $\checkmark$ & $\times$ & $\times$ & $\times$ & $\times$ & $\times$ \\

Group DRO \cite{Sagawa2020}
& $\times$ & $\times$ & $\times$ & $\times$ & $\checkmark$ & $\times$ \\
\bottomrule
\textbf{CIFA (Ours)}
& $\checkmark$ & $\checkmark$ & $\checkmark$ & $\checkmark$ & $\checkmark$ & $\checkmark$ \\
\bottomrule
\end{tabular}
}
\end{table}

As summarized in Table~\ref{tab:framework_comparison}, the distinction of CIFA lies in the explicit integration of demographic, contextual, and contextual-intersectional auditing within a common audit--mitigate--reaudit protocol. Rather than treating mitigation as the endpoint, CIFA reuses the original subgroup definitions to verify whether worst-group vulnerabilities are reduced and whether new disparities emerge after mitigation.

\section{Method}

\begin{figure}[t]
\centering
\includegraphics[width=12cm, height=7cm]{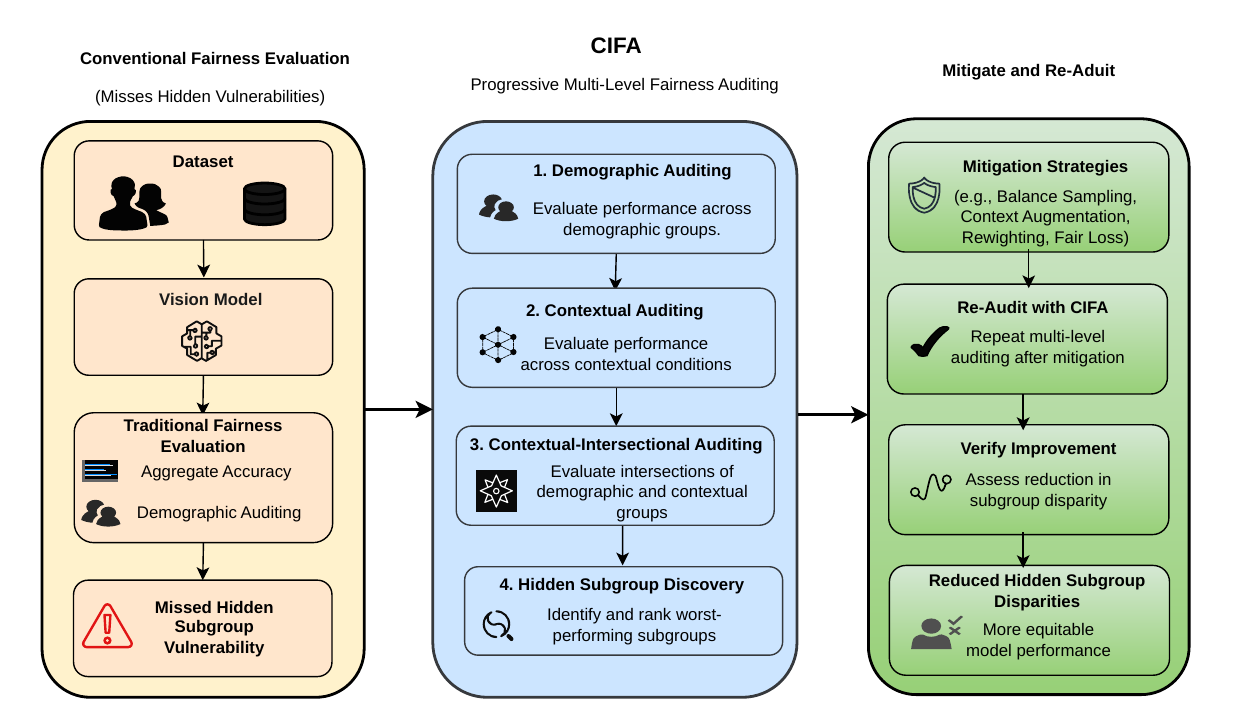}
\caption{Overview of the proposed Contextual-Intersectional Fairness Auditing Framework (CIFA). Conventional fairness evaluation primarily relies on aggregate accuracy and demographic subgroup analysis, which may overlook hidden contextual-intersectional vulnerabilities. CIFA progressively performs demographic, contextual, and contextual-intersectional auditing, identifies hidden subgroup vulnerabilities, applies mitigation, and re-applies the same auditing protocol to verify whether subgroup disparities are reduced.}
\label{fig:cifa_framework}
\end{figure}

We propose the Contextual-Intersectional Fairness Auditing Framework (CIFA), a structured audit--mitigate--re-audit protocol for identifying hidden subgroup vulnerabilities in visual models. CIFA audits performance across demographic, contextual, and contextual-intersectional groups, capturing failures that may be missed by aggregate accuracy or demographic-only evaluation. After mitigation, the same subgroup definitions are reused to measure changes in worst-group accuracy, subgroup disparity, and contextual robustness, enabling a consistent assessment of whether the identified vulnerabilities are reduced. Figure~\ref{fig:cifa_framework} presents an overview of CIFA and illustrates its audit--mitigate--re-audit pipeline.

\subsection{Contextual-Intersectional Fairness Auditing Framework}
\label{subsec:cifa_framework}

The proposed CIFA framework is a unified protocol for diagnosing hidden subgroup vulnerabilities in visual models. Unlike conventional fairness evaluation, which typically compares performance across demographic groups alone, CIFA explicitly audits both demographic attributes and visual context. The goal is to determine whether model failures arise from demographic attributes, contextual conditions, or their interactions, and to identify hidden subgroup vulnerabilities that remain undetected under conventional fairness evaluation.

Let $\mathcal{D}=\{(x_i,y_i,\mathbf{a}_i,\mathbf{c}_i)\}_{i=1}^{N}$ denote an evaluation set, where $x_i$ is an input image, $y_i$ is the target label, $\mathbf{a}_i$ denotes demographic attributes, and $\mathbf{c}_i$ denotes contextual attributes. In our setting, demographic attributes include race, gender, and age depending on the dataset, while contextual attributes comprise dataset-specific visual factors. Given a classifier $f_{\theta}$, CIFA evaluates model performance over three families of groups:
\begin{equation}
    \mathcal{G}_{d} = \{g(\mathbf{a})\}, \qquad
    \mathcal{G}_{c} = \{g(\mathbf{c})\}, \qquad
    \mathcal{G}_{dc} = \{g(\mathbf{a},\mathbf{c})\},
\end{equation}
where $\mathcal{G}_{d}$ denotes demographic groups, $\mathcal{G}_{c}$ denotes contextual groups, and $\mathcal{G}_{dc}$ denotes contextual-intersectional groups formed by combining demographic and contextual attributes. Within CIFA, a contextual-intersectional subgroup is defined as the intersection of one or more demographic attributes with one or more contextual attributes. This formulation enables the framework to evaluate model behaviour under specific contextual-intersectional combinations that cannot be observed through demographic-only or contextual-only analysis. For example, a contextual-intersectional group may be defined as: 
\begin{equation}
    g = \text{race} \times \text{age} \times \text{illumination} \times \text{image quality}.
\end{equation}
This formulation allows CIFA to move beyond aggregate and demographic-only evaluation and directly measure whether specific demographic groups are disproportionately affected under particular visual conditions.

\subsubsection{Demographic and contextual auditing:}
CIFA first evaluates model performance across demographic groups and contextual groups separately. For each group $g$, we compute the group-level accuracy
\begin{equation}
    A_g =
    \frac{1}{|\mathcal{D}_g|}
    \sum_{(x_i,y_i)\in \mathcal{D}_g}
    \mathbbm{1}\left[f_{\theta}(x_i)=y_i\right],
\end{equation}
where $\mathcal{D}_g$ is the subset of samples belonging to group $g$. Demographic auditing estimates whether performance varies across protected or demographic attributes, while contextual auditing estimates whether performance varies across visual conditions. These two stages provide complementary views of model behavior: the former identifies demographic disparities, while the latter reveals sensitivity to image acquisition and visual quality factors.

\subsubsection{Contextual-intersectional auditing:}
CIFA then evaluates groups formed by the demographic and contextual attributes. This stage is central to the framework because failures may emerge only when both dimensions are considered jointly. For each contextual-intersectional group $g \in \mathcal{G}_{dc}$, CIFA computes the same group-level metric $A_g$ and compares it against aggregate, demographic-only, and contextual-only performance. This allows the framework to detect cases where a model appears reliable at the aggregate level but fails for specific contextual-intersectional subgroups. To avoid unstable estimates from very small groups, we report worst-group statistics over reliable groups satisfying $|\mathcal{D}_g| \geq m$, where $m$ is a minimum group-size threshold.

\subsubsection{Worst-group discovery:}
After computing subgroup performance, CIFA ranks all evaluated groups according to their performance degradation. Let $A_{\mathrm{all}}$ denote the aggregate accuracy over the full evaluation set. We define the overall-to-group gap as
\begin{equation}
    \Delta_g = A_{\mathrm{all}} - A_g.
\end{equation}
The worst-group accuracy and worst-group gap are then given by
\begin{equation}
    A_{\mathrm{worst}} = \min_{g \in \mathcal{G}^{r}} A_g,
    \qquad
    \Delta_{\mathrm{worst}} = A_{\mathrm{all}} - A_{\mathrm{worst}},
\end{equation}
where $\mathcal{G}^{r}$ denotes the set of reliable groups. CIFA also reports the standard deviation of group accuracies,
\begin{equation}
    \sigma_{\mathcal{G}} =
    \sqrt{
    \frac{1}{|\mathcal{G}^{r}|}
    \sum_{g \in \mathcal{G}^{r}}
    (A_g - \bar{A}_{\mathcal{G}})^2
    },
\end{equation}
where $\bar{A}_{\mathcal{G}}$ is the mean accuracy across reliable groups. These metrics provide a compact summary of subgroup disparity, while the ranked group list identifies the specific attribute combinations responsible for the largest failures. This ranking is used to distinguish visible disparities, which are already apparent under demographic or contextual auditing, from hidden disparities that only emerge under contextual-intersectional auditing.

\subsubsection{Audit--mitigate--reaudit protocol:}
CIFA further supports a consistent audit--mitigate--reaudit protocol. In the first stage, a baseline model is audited across aggregate, demographic, contextual, and contextual-intersectional levels. The resulting subgroup rankings identify the most vulnerable groups. In the second stage, mitigation strategies are applied to reduce the identified disparities. In this work, we evaluate established mitigation strategies rather than proposing a new mitigation algorithm. These include context augmentation, label-aware balanced sampling, group-weighted cross-entropy, and Group DRO. In the final stage, the same CIFA auditing procedure is repeated after mitigation. Mitigation effectiveness is measured by comparing pre- and post-mitigation values of $A_{\mathrm{worst}}$, $\Delta_{\mathrm{worst}}$, and $\sigma_{\mathcal{G}}$ under the same group definitions.

This design avoids relying solely on aggregate accuracy, which may improve while subgroup failures persist, and evaluates mitigation against the same hidden vulnerabilities identified during auditing. CIFA therefore provides a reproducible framework for diagnosing contextual-intersectional bias, ranking worst-case subgroups, and assessing mitigation effectiveness. Importantly, CIFA is model-agnostic and can be applied to any supervised computer vision model capable of producing predictions over labelled datasets, independent of the underlying architecture or mitigation strategy.

\subsection{Mitigation Strategies} 
\label{subsec:mitigation_strategies} 

Let $\mathcal{D}_{\mathrm{train}}=\{(x_i,y_i,\mathbf{a}_i,\mathbf{c}_i)\}_{i=1}^{N}$ denote the training set, where $x_i$ is an input image, $y_i \in \mathcal{Y}$ is the gender label, $\mathbf{a}_i$ denotes demographic attributes, and $\mathbf{c}_i$ denotes contextual attributes. Each sample is assigned to a contextual-intersectional group 
\begin{equation} 
    g_i^{dc} = 
    \phi(\mathbf{a}_i,\mathbf{c}_i), 
    \qquad g_i^{dc} \in \mathcal{G}_{dc}, 
    \end{equation} 
where $\phi(\cdot)$ maps demographic and contextual attributes to a discrete subgroup. In our experiments, this group is defined as 
\begin{equation} 
g_i^{dc} = (\text{race}_i, \text{age}_i, \text{illumination}_i, \text{image quality}_i, \text{facial accessories}_i). 
\end{equation} Given a classifier $f_{\theta}$ that predicts $p_{\theta}(y_i \mid x_i)$, the baseline optimizes the average cross-entropy loss as follows: 
\begin{equation} 
\mathcal{L}(\theta) = \frac{1}{N} \sum_{i=1}^{N} -\log p_{\theta}(y_i \mid x_i). \end{equation} 

This optimizes average performance and may overlook disparities across contextual-intersectional groups. We therefore applied the following  mitigation strategies using the same group definitions as in the auditing protocol.

\subsubsection{Context Augmentation:}
We apply label preserving contextual transformations to reduce sensitivity to illumination and image quality variation. Let $\mathcal{T}$ denote the set of augmentations, including brightness/contrast changes, blur, noise, and quality degradation. The augmented objective is
\begin{equation}
    \mathcal{L}_{\mathrm{aug}}(\theta)
    =
    \frac{1}{N}
    \sum_{i=1}^{N}
    \mathbb{E}_{t\sim\mathcal{T}}
    \left[
    \ell_{\mathrm{CE}}(f_{\theta}(t(x_i)),y_i)
    \right].
\end{equation}
We model illumination perturbation as
\begin{equation}
    t_{\mathrm{ill}}(x)=\operatorname{clip}(\alpha x+\beta),
\end{equation}
where $\alpha$ and $\beta$ control contrast and brightness, respectively, and $\operatorname{clip}(\cdot)$ clamps pixel values to the valid image range.

\subsubsection{Balanced Sampling:}
To increase exposure to underrepresented contextual-intersectional groups, we use a group-balanced sampler. For group $g\in\mathcal{G}_{dc}$ with $n_g$ training samples, each sample is assigned
\begin{equation}
    s_i = \frac{1}{n_{g_i^{dc}}+\epsilon},
    \qquad
    P_{\mathrm{sampler}}(i)
    =
    \frac{s_i}{\sum_{j=1}^{N}s_j}.
\end{equation}
In the label-aware version, weights are computed over group-label cells $h_i=(g_i^{dc},y_i)$ to preserve the label distribution.

\subsubsection{Context Augmentation with Balanced Sampling:}
We also combine balanced sampling with context augmentation by sampling examples from $P_{\mathrm{sampler}}$ and then applying $t\sim\mathcal{T}$:
\begin{equation}
    \mathcal{L}_{\mathrm{aug+sampler}}(\theta)
    =
    \mathbb{E}_{i\sim P_{\mathrm{sampler}}}
    \mathbb{E}_{t\sim\mathcal{T}}
    \left[
    \ell_{\mathrm{CE}}(f_{\theta}(t(x_i)),y_i)
    \right].
\end{equation}

\subsubsection{Group-weighted Cross-Entropy:}
We reweight the loss according to contextual-intersectional group frequency. Each group receives weight
\begin{equation}
    w_g =
    \left(
    \frac{N}{|\mathcal{G}_{dc}|(n_g+\epsilon)}
    \right)^{\gamma},
\end{equation}
and the training objective becomes
\begin{equation}
    \mathcal{L}_{\mathrm{GWCE}}(\theta)
    =
    \frac{1}{N}
    \sum_{i=1}^{N}
    w_{g_i^{dc}}\ell_i(\theta).
\end{equation}

\subsubsection{Sampler-weighted Cross-Entropy:}
We further evaluate the combination of balanced sampling and group-weighted loss:
\begin{equation}
    \mathcal{L}_{\mathrm{sampler+GWCE}}(\theta)
    =
    \mathbb{E}_{i\sim P_{\mathrm{sampler}}}
    \left[
    w_{g_i^{dc}}\ell_i(\theta)
    \right].
\end{equation}

\subsubsection{Group DRO:}
Finally, we evaluate Group DRO to directly improve worst-group robustness. The objective minimizes the maximum group loss:
\begin{equation}
    \min_{\theta}
    \max_{g\in\mathcal{G}_{dc}}
    \mathcal{L}_g(\theta),
    \qquad
    \mathcal{L}_g(\theta)
    =
    \frac{1}{n_g}
    \sum_{i:g_i^{dc}=g}
    \ell_i(\theta).
\end{equation}
In practice, we use adaptive group weights:
\begin{equation}
    \mathcal{L}_{\mathrm{DRO}}(\theta)
    =
    \sum_{g\in\mathcal{G}_{dc}}
    q_g\mathcal{L}_g(\theta),
    \qquad
    q_g \leftarrow
    \frac{q_g\exp(\eta\mathcal{L}_g)}
    {\sum_{g'\in\mathcal{G}_{dc}}q_{g'}\exp(\eta\mathcal{L}_{g'})}.
\end{equation}

Overall, these strategies address different possible sources of contextual-intersectional disparity: context augmentation targets sensitivity to visual conditions, balanced sampling addresses group imbalance, group-weighted loss changes group-level optimization pressure, and Group DRO explicitly emphasizes the worst-performing groups.

\section{Experiments}

\subsection{Dataset}

We evaluate CIFA on three publicly available face datasets: FairFace \cite{Karkkainen2021}, CelebA \cite{Liu2015}, and UTKFace \cite{Zhang2017}. FairFace provides balanced race, gender, and age annotations; UTKFace provides additional variation in age, gender, and ethnicity; and CelebA provides rich facial attribute annotations for contextual auditing.

For each dataset, we construct demographic and contextual group labels according to the available annotations. Contextual attributes are obtained either directly from dataset annotations or derived from image-level measurements. For CelebA, annotated facial attributes such as eyeglasses, facial hair, makeup, and accessories are used as contextual variables. For datasets without explicit contextual annotations, we estimate illumination from image brightness and image quality from the variance of the Laplacian. Continuous contextual measurements are discretized into low, medium, and high groups using quantile-based binning, enabling consistent contextual and contextual-intersectional auditing across datasets.

\subsection{Evaluation Metric} 

For every demographic, contextual, and contextual-intersectional subgroup, CIFA computes subgroup accuracy. False-positive rate (FPR) and false-negative rate (FNR) are additionally computed during auditing to characterize subgroup error behaviour. In this paper, the comparative analysis focuses on accuracy-based fairness measures. Accordingly, we report overall accuracy, worst-group accuracy, the overall-to-worst-group accuracy gap, and the standard deviation of subgroup accuracies.

\subsection{Implementation Details}
\label{subsec:implementation_details}

We evaluate CIFA on gender classification using FairFace \cite{Karkkainen2021}, CelebA \cite{Liu2015}, and UTKFace \cite{Zhang2017} datasets. All input images are resized to $224 \times 224$ and normalized using ImageNet mean and standard deviation \cite{deng2009imagenet}. We use ResNet-50 \cite{he2016deep} and ViT-B/16 \cite{dosovitskiy2020image} as baseline architectures, both initialized with ImageNet-pretrained weights and fine-tuned with a two-class classification head. Models are trained for 10 epochs using Adam optimizer \cite{kingma2014adam}. ResNet-50 is trained with batch size $64$ and learning rate $1\mathrm{e}{-4}$, while ViT-B/16 is trained with learning rate $3\mathrm{e}{-5}$ and weight decay $0.05$ with a batch size of $32$.

For mitigation, we evaluate baseline, context augmentation, label-aware balanced sampling, group-weighted cross-entropy, Group DRO, and their sampler based combinations using the same contextual-intersectional group definitions as the CIFA audit. Context augmentation applies label preserving brightness or contrast perturbation, saturation and hue jittering, Gaussian blur, noise, and image quality degradation. Brightness and contrast jitter are set to $0.25$, saturation to $0.10$, hue to $0.02$, and Gaussian blur is applied with probability $0.20$. Sampling and loss-weighting methods use inverse square-root group frequency, with group weights normalized to mean one and clipped at $3.0$. Group DRO updates group weights with step size $\eta=0.02$. All experiments are implemented in PyTorch and run on NVIDIA Tesla V100 GPUs with a fixed random seed of $42$.

\subsection{Demographic and Contextual Auditing}
\label{subsec:contextual_auditing}
CIFA first evaluates demographic and contextual factors independently before constructing contextual-intersectional groups. Demographic auditing measures subgroup performance across attributes such as age, race, and gender, while contextual auditing applies the same protocol to dataset-specific visual factors. For FairFace \cite{Karkkainen2021} and UTKFace \cite{Zhang2017}, contextual groups are defined using illumination and image-quality categories; for CelebA \cite{Liu2015}, they are defined using binary attributes such as blur, eyeglasses, hats, facial hair, and makeup. For each subgroup, CIFA computes accuracy, false-positive rate (FPR), and false-negative rate (FNR) using the subgroup definitions and evaluation protocol described in Section~\ref{subsec:cifa_framework}.

\begin{table}[h]
\centering
\caption{Summary of the lowest-performing contextual subgroup identified during contextual auditing. Contextual auditing evaluates each contextual attribute independently by computing subgroup accuracy, false-positive rate (FPR), and false-negative rate (FNR) before demographic and contextual attributes are combined during contextual-intersectional auditing. For brevity, only subgroup accuracy is reported here, while FPR and FNR are computed during the audit to characterise subgroup error behaviour.}
\label{tab:contextual_audit}
\small
\setlength{\tabcolsep}{5pt}
\begin{tabular}{cccc}
\toprule
\textbf{Model} &
\textbf{Dataset} &
\textbf{Worst Contextual Subgroup} &
\textbf{Acc. (\%)} \\
\midrule

\multirow{3}{*}{ResNet-50}
& FairFace \cite{Karkkainen2021} & Low Image Quality & 83.38 \\
& UTKFace \cite{Zhang2017}  & High Illumination & 86.59 \\
& CelebA \cite{Liu2015}   & Blurry            & 89.31 \\
\midrule

\multirow{3}{*}{ViT-B/16}
& FairFace \cite{Karkkainen2021} & Low Image Quality & 87.91 \\
& UTKFace \cite{Zhang2017} & High Illumination & 90.66 \\
& CelebA \cite{Liu2015}  & Eyeglasses        & 92.79 \\
\bottomrule
\end{tabular}
\end{table}

In Table~\ref{tab:top_contextual_intersectional}, the overall accuracy reflects demographic performance, while Table~\ref{tab:contextual_audit} reports the lowest-performing contextual subgroup for each dataset and model, identified from the complete set of contextual audit results. Image quality and illumination cause the largest degradation on FairFace \cite{Karkkainen2021} and UTKFace \cite{Zhang2017}, while blur and eyeglasses are most influential on CelebA \cite{Liu2015}. These findings demonstrate that visual context independently influences model performance. However, contextual auditing alone cannot determine whether these effects disproportionately affect particular demographic groups, motivating the subsequent contextual-intersectional audit.

\subsection{Contextual-Intersectional Vulnerabilities}
Table~\ref{tab:top_contextual_intersectional} summarizes the worst contextual-intersectional subgroup performance across FairFace \cite{Karkkainen2021}, UTKFace \cite{Zhang2017}, and CelebA \cite{Liu2015}. Although overall accuracy remains high, ranging from $92.34\%$ to $98.94\%$, substantially lower performance is observed for the worst subgroups. The largest disparity occurs on UTKFace with ResNet-50, where accuracy drops from $92.34\%$ overall to $65.91\%$ for the worst subgroup, yielding a gap of $26.43\%$. FairFace also exhibits gaps above $21\%$ for both architectures, while CelebA shows smaller but persistent disparities of $7.51\%$ and $8.17\%$. ViT-B/16 improves worst-group performance on UTKFace but does not consistently reduce disparities across all datasets. These results show that strong aggregate accuracy can mask substantial contextual-intersectional failures, motivating targeted mitigation followed by systematic re-auditing.

\begin{table}[]
\centering
\caption{Worst contextual-intersectional subgroup performance across datasets and model architectures. The gap is computed as overall accuracy minus worst-group accuracy. All values are reported in \%.}
\label{tab:top_contextual_intersectional}
\small
\setlength{\tabcolsep}{5pt}
\resizebox{\textwidth}{!}{
\begin{tabular}{llcccc}
\toprule
\textbf{Model} &
\textbf{Dataset} &
\textbf{Overall Acc.} &
\textbf{Worst-group Acc.} &
\textbf{Gap} &
\textbf{Group Std.} \\
\midrule

\multirow{3}{*}{ResNet-50}
& FairFace \cite{Karkkainen2021} & 92.78 & 71.74 & 21.04 & 6.09 \\
& UTKFace \cite{Zhang2017} & 92.34 & 65.91 & 26.43 & 7.97 \\
& CelebA \cite{Liu2015}  & 98.94 & 91.43 & 7.51  & 2.71 \\
\midrule

\multirow{3}{*}{ViT-B/16}
& FairFace \cite{Karkkainen2021} & 93.04 & 71.43 & 21.61 & 5.78 \\
& UTKFace \cite{Zhang2017} & 93.67 & 74.19 & 19.48 & 6.38 \\
& CelebA \cite{Liu2015}  & 98.73 & 90.57 & 8.17  & 2.67 \\
\bottomrule
\end{tabular}
}
\end{table}

\subsection{Contextual-Intersectional Mitigation and Re-Auditing}
\label{sec:mitigation_reaudit}

Table~\ref{tab:resnet50_mitigation_results} and Table~\ref{tab:vitb16_mitigation_results} report the mitigation results after re-applying the CIFA audit protocol to each trained model. Rather than evaluating mitigation only by aggregate accuracy, we use the same contextual-intersectional group definitions to measure changes in worst-group accuracy, overall-to-worst-group gap, and group-level variance. Across both ResNet-50 and ViT-B/16, several mitigation strategies reduce the subgroup vulnerabilities identified in the initial audit. For ResNet-50, group-weighted cross-entropy improves the worst-group accuracy on UTKFace from $65.91\%$ to $74.19\%$ and reduces group variance, while Group DRO yields the largest gap reduction, decreasing the gap from $26.43\%$ to $16.68\%$. On FairFace, context augmentation and group-weighted cross-entropy both improve worst-group accuracy from $71.74\%$ to $74.29\%$. On CelebA, group-weighted cross-entropy performs best, increasing worst-group accuracy from $91.43\%$ to $92.45\%$ and reducing the gap from $7.51\%$ to $6.55\%$.

\begin{table*}[h]
\centering
\caption{Mitigation and re-audit results using ResNet-50 on UTKFace, FairFace, and CelebA. \textbf{OA} and \textbf{WGA} denote overall and worst-group accuracy, respectively; Gap is their difference, and Std. is the standard deviation of subgroup accuracies. Metrics are reported over reliable contextual-intersectional groups. All values are reported in \%.}
\label{tab:resnet50_mitigation_results}
\small
\setlength{\tabcolsep}{5pt}
\resizebox{\textwidth}{!}{
\begin{tabular}{lcccccccccccc}
\toprule
\textbf{Method} &
\multicolumn{4}{c}{\textbf{UTKFace} \cite{Zhang2017}} &
\multicolumn{4}{c}{\textbf{FairFace} \cite{Karkkainen2021}} &
\multicolumn{4}{c}{\textbf{CelebA} \cite{Liu2015}} \\
\cmidrule(lr){2-5}
\cmidrule(lr){6-9}
\cmidrule(lr){10-13}
&
\textbf{OA} &
\textbf{WGA} &
\textbf{Gap} &
\textbf{Std.} &
\textbf{OA} &
\textbf{WGA} &
\textbf{Gap} &
\textbf{Std.} &
\textbf{OA} &
\textbf{WGA} &
\textbf{Gap} &
\textbf{Std.} \\
\midrule

Baseline
& 92.34 & 65.91 & 26.43 & 7.97
& 92.78 & 71.74 & 21.04 & 6.09
& 98.94 & 91.43 & 7.51 & 2.71 \\

Balanced Sampling
& 92.36 & 69.05 & 23.32 & 7.57
& 92.24 & 71.74 & 20.50 & 6.48
& 98.67 & 91.43 & 7.24 & 2.59 \\

Context Aug.
& 93.21 & 70.97 & 22.24 & 7.27
& 92.70 & \textbf{74.29} & 18.41 & 5.80
& 99.04 & 91.43 & 7.62 & 2.71 \\

Context Aug. + Sampling
& 93.17 & 70.97 & 22.20 & 7.34
& 92.72 & 68.57 & 24.15 & 6.12
& 98.94 & 91.43 & 7.51 & 2.56 \\

Sampling + Weighted CE
& 92.13 & 70.45 & 21.68 & 7.54
& 91.63 & 70.97 & 20.66 & 6.78
& 98.86 & 88.68 & 10.18 & 3.21 \\

Group DRO
& 89.41 & 72.73 & \textbf{16.68} & 7.28
& 89.96 & 70.97 & 18.99 & 6.44
& 98.67 & 89.62 & 9.04 & 2.69 \\

Group-weighted CE
& 92.70 & \textbf{74.19} & 18.51 & 6.05
& 92.52 & \textbf{74.29} & \textbf{18.24} & 5.77
& 99.00 & \textbf{92.45} & \textbf{6.55} & 2.13 \\

\bottomrule
\end{tabular}
}
\end{table*}

For ViT-B/16, the most effective mitigation strategy varies across datasets. On FairFace, context augmentation improves worst-group accuracy from $71.43\%$ to $77.42\%$, while on UTKFace, Group DRO increases it from $74.19\%$ to $78.57\%$ and reduces the gap from $19.48$ to $14.89$ percentage points. On CelebA, sampling with weighted cross-entropy performs best, raising worst-group accuracy from $90.57\%$ to $92.45\%$ and reducing the gap from $8.17$ to $6.20$ points. These results show that no single mitigation strategy consistently dominates across datasets and architectures. Re-auditing is therefore essential to verify improvements in the most vulnerable contextual-intersectional groups rather than aggregate accuracy alone.

\begin{table*}[h]
\centering
\caption{Mitigation and re-audit results using ViT-B/16 on UTKFace, FairFace, and CelebA. \textbf{OA} and \textbf{WGA} denote overall and worst-group accuracy, respectively; Gap is their difference, and Std. is the standard deviation of subgroup accuracies. Metrics are reported over reliable contextual-intersectional groups. All values are reported in \%.}
\label{tab:vitb16_mitigation_results}
\small
\setlength{\tabcolsep}{5pt}
\resizebox{\textwidth}{!}{
\begin{tabular}{lcccccccccccc}
\toprule
\textbf{Method} &
\multicolumn{4}{c}{\textbf{UTKFace} \cite{Zhang2017}} &
\multicolumn{4}{c}{\textbf{FairFace} \cite{Karkkainen2021}} &
\multicolumn{4}{c}{\textbf{CelebA} \cite{Liu2015}} \\
\cmidrule(lr){2-5}
\cmidrule(lr){6-9}
\cmidrule(lr){10-13}
&
\textbf{OA} &
\textbf{WGA} &
\textbf{Gap} &
\textbf{Std.} &
\textbf{OA} &
\textbf{WGA} &
\textbf{Gap} &
\textbf{Std.} &
\textbf{OA} &
\textbf{WGA} &
\textbf{Gap} &
\textbf{Std.} \\
\midrule

Baseline
& 93.67 & 74.19 & 19.48 & 6.38
& 93.04 & 71.43 & 21.61 & 5.78
& 98.73 & 90.57 & 8.17 & 2.67 \\

Balanced Sampling
& 93.00 & 76.71 & 16.28 & 5.96
& 92.22 & 74.07 & 18.15 & 6.55
& 98.84 & 90.57 & 8.27 & 2.60 \\

Context Aug.
& 92.39 & 71.43 & 20.96 & 6.45
& 93.12 & \textbf{77.42} & 15.70 & 5.44
& 98.22 & 89.58 & 8.64 & 2.97 \\

Context Aug.
+ Sampling
& 93.10 & 76.19 & 16.91 & 5.95
& 92.67 & 72.73 & 19.94 & 6.20
& 98.48 & 89.58 & 8.90 & 2.92 \\

Sampling + Weighted CE
& 93.12 & 72.73 & 20.40 & 6.29
& 92.24 & 73.33 & 18.91 & 5.96
& 98.66 & \textbf{92.45} & \textbf{6.20} & 2.13 \\

Group DRO
& 93.46 & \textbf{78.57} & \textbf{14.89} & 5.57
& 92.05 & 74.19 & 17.86 & 5.85
& 98.68 & 88.57 & 10.10 & 3.05 \\

Group-weighted CE
& 93.57 & 75.00 & 18.57 & 5.40
& 92.82 & 71.43 & 21.40 & 6.16
& 98.81 & 91.11 & 7.70 & 2.78 \\

\bottomrule
\end{tabular}
}
\end{table*}

\subsection{Limitations and Discussion}
\label{sec:Limitations}
This study evaluates CIFA on gender classification using face datasets that provide demographic annotations and contextual or appearance-related attributes suitable for systematic auditing. The results therefore should not be interpreted as exhaustive evidence across all computer vision tasks, model families, or deployment settings. In addition, some contextual attributes are estimated from image-level statistics rather than human annotations. While this improves reproducibility, it may not capture all semantically meaningful forms of visual context. Future work should extend CIFA to broader tasks, datasets, foundation models, and deployment-specific contextual factors.

\section{Conclusion}

We introduced CIFA, a contextual-intersectional fairness auditing framework for identifying hidden subgroup vulnerabilities in face analysis systems. By jointly evaluating demographic, contextual, and contextual-intersectional groups, CIFA reveals performance disparities that are not captured by aggregate accuracy or demographic-only analysis. Experiments on FairFace, CelebA, and UTKFace with ResNet-50 and ViT-B/16 show that strong overall performance can coexist with substantial worst-group failures, underscoring the need to account for the interaction between demographic characteristics and visual context. We further evaluated CIFA within an audit--mitigate--reaudit workflow. Standard mitigation strategies improved several worst-group outcomes, but their effectiveness varied across datasets and architectures, and no single method consistently removed all contextual-intersectional disparities. These results show that mitigation should be followed by systematic re-auditing rather than assessed through aggregate performance alone. Overall, CIFA provides a reproducible framework for discovering, prioritizing, and reassessing hidden subgroup risks, and offers a practical foundation for more comprehensive fairness evaluation of visual models.

\newpage
\section*{Acknowledgements}
This work is supported by the Villum Synergy Grant No. 57384, titled XAI for Safety and Security: A Bottom-Up Approach.

%
%
\bibliographystyle{splncs04}
\bibliography{main}
\end{document}